# A Quantum Neural Network Transfer-Learning Model for Forecasting Problems with Continuous and Discrete Variables

Ismael Abdulrahman, ismael.abdulrahman@epu.edu.iq
Department of Technical Information Systems Engineering, Erbil Technical Engineering College, Erbil Polytechnic University, Erbil 44001, Kurdistan region–Iraq.

*Abstract*— This study introduces simple yet effective continuous- and discrete-variable quantum neural network (QNN) models as a transfer-learning approach for forecasting tasks. The CV-QNN features a single quantum layer with two qubits to establish entanglement and utilizes a minimal set of quantum gates, including displacement, rotation, beam splitter, squeezing, and a non-Gaussian cubic-phase gate, with a maximum of eight trainable parameters. A key advantage of this model is its ability to be trained on a single dataset, after which the learned parameters can be transferred to other forecasting problems with little to no fine-tuning. Initially trained on the Kurdistan load demand dataset, the model's frozen parameters are successfully applied to various forecasting tasks, including energy consumption, traffic flow, weather conditions, and cryptocurrency price prediction, demonstrating strong performance. Furthermore, the study introduces a discrete-variable quantum model with an equivalent 2- and 4-wire configuration and presents a performance assessment, showing good but relatively lower effectiveness compared to the continuous-variable model.

**1.1 Introduction**
Quantum computers in the near term are available as either superconducting quantum computers, which use the discrete variable (qubit-based) model, or photonic quantum computers, which operate on the continuous variable model. Superconducting computers utilize the particle-like properties of nature, employing individual particles such as electrons as information carriers. These particles can exist in two states, up or down, corresponding to "0" and "1", making the computational states finite in the qubit model. To maintain the stability of these states, extremely low temperatures are required, necessitating that the chips be housed in dilution systems. Each qubit is connected to classical bit hardware via wires, and sophisticated engineering is needed to control the wire temperature to avoid disturbing the qubit states [1-4].

In contrast, photonic quantum computers use bosonic modes (qumodes) as information carriers, which can be generated and sustained at room temperature using linear optical devices. This makes them easily integrable into existing computing systems. The computational state space of the continuous variable (CV) model, which photonic computers are based on, is infinite-dimensional and allows for a broader range of quantum gates compared to the qubit model [2].

Research on implementing machine learning algorithms on quantum computers is ongoing. Quantum machine learning algorithms can be realized using variational quantum circuits with parametrized quantum gates. A quantum circuit consists of quantum gates, and the change in the initial quantum state caused by the circuit represents quantum computation. The results of quantum computing are obtained through measurement and integrated into optimization and parameter updates on classical circuits [5].

Quantum neural networks (QNNs), a subset of quantum machine learning, operate similarly to classical neural networks. In QNNs, the quantum processing unit (QPU) handles the network operations, while optimization takes place on classical processors. Classical neural networks rely on two main components: linear transformations and nonlinear activation functions. In the qubit-based model, all unitary gates are linear, which makes it difficult to directly implement bias addition and nonlinear activation functions. However, in the continuous variable model, the displacement gate can handle bias addition, and the Kerr and cubic-phase gate provides the nonlinear activation function. This enables a natural adaptation of the classical neural network architecture into the quantum framework [5].

The concepts, applications, and challenges of quantum neural networks (QNNs) are presented in [6]. The training process for QNNs is covered in [7], while [8] explores their potential capabilities. Deep QNNs are addressed in [9], with [10] focusing on implementations specifically using superconducting processors. Quantum optimal neural networks are detailed in [11], and quantum convolutional neural networks (CNNs) are explored in [12]. Multi-qubit QNNs are examined in [13]. A concise review of QNN methods and applications is provided in [14], and a scalable, noise-resilient QNN for noisy intermediate-scale quantum (NISQ) computers is presented in [15]. Additionally, the optimization of QNNs is discussed in [16], a co-design framework for QNNs is outlined in [17], and quantum embedding methods with transformer QNNs for strongly correlated materials are highlighted in [18].

The highlighted studies primarily focus on quantum neural networks based on the qubit model, also known as the discrete variable model. In contrast, the literature reveals only a limited number of studies addressing continuous variable quantum photonic computers. In [19], the adaptability of continuous-variable quantum neural networks is demonstrated through experiments with the Strawberry Fields software library, showcasing applications in fraud detection and image generation. The study proposes a method for building neural networks on quantum computers using a variational quantum circuit in the CV architecture, where the sizes of the later layers are progressively reduced. In [20], Physics-Informed Neural Networks (PINNs) are implemented using a continuous variable quantum framework to tackle the one-dimensional Poisson problem. In [21], an efficient cryptography scheme based on a CV quantum neural network (CV-QNN) is designed for key generation, encryption, and decryption processes on the Strawberry Fields platform. This scheme shows promise for practical applications in quantum devices. In [22], four quantum models are proposed for the CV model using a combination of Gaussian gates (displacement, rotation, squeezing, beamsplitter) and non-Gaussian gates such as the Kerr gate. In [23-24], a detailed introduction to photonic quantum computers is provided.

This study introduces a simple yet powerful CV-QNN model featuring two wires and eight tunable parameters at the most, utilizing seven quantum gates with minimal resources for forecasting problems. The pretrained model is first trained on a load demand forecasting dataset and then transferred, with its parameters frozen, to other similar problems, including energy consumption prediction, traffic flow estimation, weather condition forecasting, and cryptocurrency price prediction. Additionally, two versions of the DV-QNN are developed using qubit-based gates, with configurations of two and four wires. The study employs several different forecasting datasets, and the results are compared with those from recent studies.

**1.2 Continuous-Variable Quantum Neural Networks**
The basic unit of information in continuous variable quantum computing is qumode which is denoted as $|\psi\rangle$ and can be expressed using a quantum state basis expansion as follows [20]:
$$|\psi\rangle = \int \psi(x) |x\rangle \, dx$$
where the states $|x\rangle$ are eigenstates of the $\hat{x}$ quadrature, with the eigenvalue $x$ being real-valued. This approach is different from the qubit-based system, where the qubit $|\phi\rangle$ is defined as a superposition of the states $|0\rangle$ and $|1\rangle$:
$$|\phi\rangle = \phi_0 |0\rangle + \phi_1 |1\rangle$$
Unlike qubit-based quantum computing, where the coefficients are discrete, the CV model operates on a continuum of coefficients (continuous eigenvalue spectrum). This distinction underpins the name of this approach. The position ($\hat{x}$) and momentum ($\hat{p}$) operators, which define phase space, are examples of continuous quantum operators and play a key role in CV quantum computing. The position operator is defined as:
$$\hat{x} = \int_{-\infty}^{\infty} x |x\rangle \langle x| \, dx$$
where the vectors $|x\rangle$ are orthogonal. Similarly, the momentum operator is defined as:
$$\hat{p} = \int_{-\infty}^{\infty} p |p\rangle \langle p| \, dp$$
where $|p\rangle$ are also orthogonal vectors. A qumode is related to a pair of position and momentum operators, $(\hat{x}, \hat{p})$, which do not commute, resulting in the Heisenberg uncertainty principle for simultaneous measurements of $\hat{x}$ and $\hat{p}$. Similar to qubit-based quantum computing, CV quantum computation can also be described using fundamental gates that can be implemented through optical devices. CV quantum programming involves sequences of these gates operating on qumodes. Four fundamental Gaussian gates are known for constructing CV quantum neural networks:

1. **Displacement Gate - $D(\alpha)$**: this gate shifts the phase space by a complex number $\alpha$.
$$D(\alpha): \begin{bmatrix} x \\ p \end{bmatrix} \to \begin{bmatrix} x + \Re(\alpha) \\ p + \Im(\alpha) \end{bmatrix}$$

2. **Rotation Gate $R(\phi)$:** this operation rotates the phase space by an angle $\phi$.
$$R(\phi): \begin{bmatrix} x \\ p \end{bmatrix} \to \begin{bmatrix} cos\phi & sin\phi \\ -sin\phi & cos\phi \end{bmatrix} \begin{bmatrix} x \\ p \end{bmatrix}$$

3. **Squeezing Gate $S(r)$:** this gate scales the phase space with a factor $r$.
$$S(r): \begin{bmatrix} x \\ p \end{bmatrix} \to \begin{bmatrix} e^{-r} & 0 \\ 0 & e^{-r} \end{bmatrix} \begin{bmatrix} x \\ p \end{bmatrix}$$

4. **Beam-Splitter Gate $BS(\theta)$:** this operation acts as a rotation between two qumodes.
$$BS(\theta): \begin{bmatrix} x_1 \\ x_2 \\ p_1 \\ p_1 \end{bmatrix} \to \begin{bmatrix} cos\theta & -sin\theta & 0 & 0 \\ sin\theta & cos\theta & 0 & 0 \\ 0 & 0 & cos\theta & -sin\theta \\ 0 & 0 & sin\theta & cos\theta \end{bmatrix} \begin{bmatrix} x_1 \\ x_2 \\ p_1 \\ p_1 \end{bmatrix}$$

The first three gates are classified as Gaussian gates which are notable for their ability to preserve the Gaussian nature of quantum states. When applied to a Gaussian state, the output remains a Gaussian state. A key gate derived from other components is the *interferometer*, which can be represented as a combination of beam-splitter and rotation gates. When applied to a single qumode, the interferometer simplifies to a rotation gate. By integrating these Gaussian gates, we can establish an affine transformation, which plays a crucial role in the formulation of neural network computations. In addition, non-Gaussian gates, such as the Kerr or cubic gates, introduce non-linearities analogous to activation functions in classical neural networks. The Kerr gate, denoted as $K(\kappa)$, plays a significant role due to its availability in quantum simulators.

At the core of quantum computing operations is measurement. In this work, the expected value of the quadrature operator $\hat{x}$ is evaluated after the quantum computations: $\langle\psi_x | \hat{x} |\psi_x\rangle$. A promising approach for implementing CV quantum gates is through photonic technology. An example of such a quantum photonic computer is Xanadu's Borealis quantum computer, which is designed specifically to address Gaussian Boson Sampling (GBS) problems. The field of developing CV quantum computing systems is currently an area of very active research and development, [20, 25-26]. With an understanding of CV quantum gates, their parameters, and the expected value of the quadrature operator, we can proceed to develop a quantum neural network, building upon the foundational work of [20]. The core element of a quantum neural network is the quantum neural network unit, often referred to as a quantum network layer in existing literature, which functions similarly to a classical neural network unit. Figure 1 illustrates the essential components of a quantum neural unit.

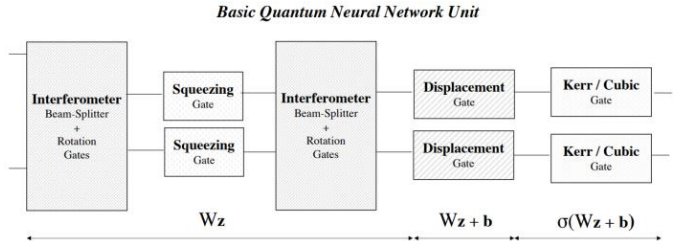

Fig. 1 Components of a quantum neural unit [20]

The first three components consist of a sequence of operations: a first interferometer, a squeezing gate, and a second interferometer. These operations result in an outcome that is comparable to multiplying the phase space vector by the neural network weights $W$ (which correspond to the parameters of the interferometers and squeezing gates). As in classical neural networks, a displacement gate simulates the addition of a bias $b$. Finally, a Kerr gate (or cubic gate) introduces a non-linearity, analogous to the activation function σ in classical neural networks:

$$| x\rangle \rightarrow | \sigma(Wx + b)\rangle$$

A quantum neural network can be constructed by stacking multiple quantum neural units in a sequential manner. It's crucial to recognize that for each qumode, every gate can be parameterized by seven variables ($\alpha, \phi, r, \theta, and\ \kappa$), which can either be real values ($\phi, r, \theta, and\ \kappa$) or complex numbers ($\alpha, r$). The complex parameters can be expressed as two real numbers using Cartesian coordinates (real and imaginary parts) or polar coordinates (amplitude and phase). The quantum circuit parameters can be categorized into passive and active parameters: the beam splitter angles and gate phases are passive parameters, while the displacement, squeezing, and Kerr magnitude are considered active parameters. The training process for quantum neural networks focuses on optimizing the parameter values ($\alpha, \phi, r, \theta, and\ \kappa$) across different qumodes and quantum neural units to minimize the cost function [20]. The number of parameters that these gates can handle for m-qumode circuits is 8m- 2, as illustrated [5].

**The proposed CV-QNN layer architecture**
A continuous-variable quantum neural network can be built using a single wire, but such a structure cannot generate entanglement, a core component of any quantum computing system (see [27] for instance). Thus, a two-wire QNN is generally more robust and efficient. However, the network shown in Fig. 1 is complex and requires more trainable parameters with identical gates per wire. Here, a simpler network is explored using basic gates of a CV photonic quantum computer, such as beam splitters, rotations, displacements, squeezers, and nonlinear gates like the Kerr and cubic-phase gates. It is generalized to work with any number of input features, not only two wires.

The first proposed approach to simplify the network is to "sandwich" squeezing gates between rotation gates, alternating each cycle between a positive and negative direction. The Grover diffusion operator also uses such sandwiching, though here we still require rotation gates to allow the optimizer to find the optimal rotation angles. Therefore, two rotation gates are implemented: one with a fixed quarter-cycle angle and another as an adjustable parameter for optimization. Applying this setup to both wires would increase complexity. Instead, the first proposed model applies the rotation sandwiching only on the first wire, while the second wire uses a displacement gate with fixed parameters, such as unity and a quarter-cycle angle. Both wires are connected to each other by two gates (beam splitter and the Cross-Kerr gate) that enable entanglement and nonlinearity, allowing information sharing across the wires. For this configuration, the network consists of 10 quantum gates with 8 tunable parameters but it can be reduced further to a lower number of gates and parameters.

In the second simplification of the model, the concept of Hong–Ou–Mandel (HOM) effect can be beneficial for the beamsplitter when applied to two input wires. Since the beamsplitter gate operates on two qubits (or optical modes), the HOM effect plays a fundamental role in producing coherent interference patterns. According to this effect [28], when two indistinguishable photons enter a balanced beamsplitter simultaneously, they undergo quantum interference, resulting in either a fully coherent output or complete destructive interference, leading to a zero-output signal. This phenomenon is particularly useful for generating entanglement and dynamically modulating the weight of a given sample. When interference is strong or constructive, the sample receives a higher weighting, whereas destructive interference diminishes its contribution. This enables more precise quantum state manipulation and improved decision-making in quantum algorithms. Consequently, this proposed network relies primarily on the first wire for computations. However, the second wire remains essential for the beamsplitter to dynamically adjust the sample weights based on the HOM effect. This role is analogous to an ancilla qubit in quantum computing, which is introduced to enable entanglement, perform intermediate computations, or assist in measurement-based operations without contributing directly to the final output. This approach requires only six quantum gates with six tunable parameters. It can be reduced to only four tunable parameters if the beamsplitter splits the inputs into a fifty-fifty or any other ratio by fixing the angles of the gate.

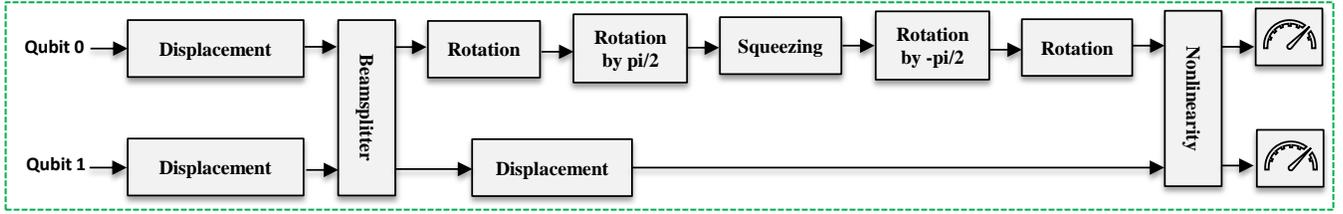

Fig. 2a The proposed two-wire CV-QNN architecture (Simplification 1)

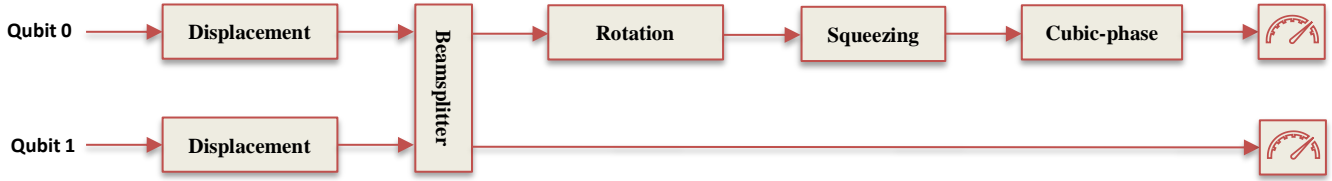

Fig. 2b The proposed two-wire CV-QNN architecture (Simplification 2) used to generate the transfer-learning weights for other datasets. The second wire is used as an auxiliary or ancilla wire used in quantum computing.

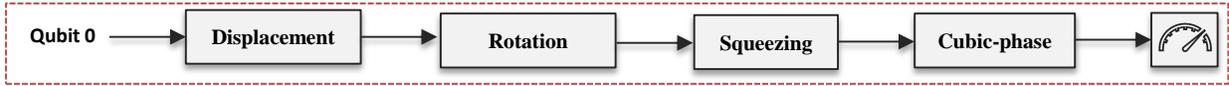

Fig. 2c The proposed two-wire CV-QNN architecture (Simplification 3, a special case for a single-feature input)

The third simplification serves as a special case of the second simplification model and there is no entanglement step in this model. If both inputs are identical, the Hong–Ou–Mandel (HOM) effect dictates that quantum interference occurs in such a way that the beamsplitter becomes redundant. As a result, the beamsplitter can be removed, allowing the system to function with only a single wire, without affecting the overall behavior of the quantum process.

Since forecasting problems can be approached by using the data as input and its shifted samples by one-time step as the output, we can apply the second simplification, while adapting the two-wire model to incorporate entanglement and additional input features. If more features are introduced, the same procedure can be extended between the first wire and each of the new features, ensuring consistency in the quantum interference process. If only one input feature is used, the beamsplitter will utilize the HOM effect to make one wire coherent while setting the other wire to zero. Therefore, the 2-wire model is more general for any number of input features. The three proposed simplifications are shown in Fig. 2.

### 1.3 Datasets Descriptions

To verify the effectiveness of the proposed QNN in solving forecasting problems, the load forecasting dataset for the Kurdistan regional network is used, as collected by the study [29]. This dataset spans six years (2015–2020), with five years allocated for training and the final year reserved for testing. Additionally, seven publicly-available forecasting datasets are employed in this study including Hourly Electricity Consumption and Production containing 46,011 samples [30], Hourly Energy Consumption by American Electric Power (AEP) containing 121,273 samples [31], Traffic Vehicle Prediction at three traffic junctions containing 48,120 samples [32], Weather Prediction containing 96,453 samples [33], the sunspot number dataset contains 75,664 samples [34], and the cryptocurrency price analysis dataset consists of five different sub-datasets [35].

**1.4 Training the Three Proposed Models**

The first dataset, which contains load demand data for Kurdistan, is employed for training the proposed models. After training, the weights are saved and utilized for transfer learning, which is then applied to the other forecasting datasets. The study evaluates several scenarios, using either a single feature (load demand in MW), two features (load demand and weather), or three features (min, max, and mean of demand), all implemented with the proposed model. For quantum simulations, this study uses PennyLane by Xanadu—an open-source, Python-based library supporting both discrete and continuous-variable quantum computations. A quantum device is defined using Strawberry Fields with a Fock backend. The experiment runs on a standard computer with 16 GB RAM, a 2.4 GHz processor.

The classical data is mapped to quantum states using displacement operators, which encode the classical input into quantum states by shifting them in phase space. Specifically, the classical values are fed into the quantum system using displacement encoding, where the displacement operator is applied to each classical feature. This operation shifts the quantum state in phase space, representing the classical data in terms of its position and momentum.

For the optimization process, the loss function used is square loss, which computes the mean squared error between the predicted and actual labels. The cost function is designed to optimize the quantum neural network, utilizing the square loss function to measure the discrepancy between the model's predictions and the true labels. The training of the quantum model is done using the Adam optimizer, with an initial learning rate of 0.005 and beta values of 0.9 and 0.999. The optimizer adjusts the learning rate dynamically during training based on the error percentage. Specifically, if the error is below certain thresholds, the learning rate is reduced progressively to fine-tune the training process. The quantum neural network model is initialized with a cut-off dimension of 12 Fock states, and the initial values for the model parameters are randomly set using a Python seed of zero. The number of epochs is set to 20/50/100 epochs for the three models, respectively. The model structures with corresponding optimized parameter values are shown in Fig. 3. The diagrams represent a single-layer QNN. While it is possible to stack multiple such layers for a deeper network, this requires high computational resources. Therefore, the proposed QNN will use only single layer network. The learning process (cost vs. epoch) for the models is shown in Fig. 4. The predicted load demand for the test dataset (sixth year), compared to the actual values, is shown in Fig. 5.

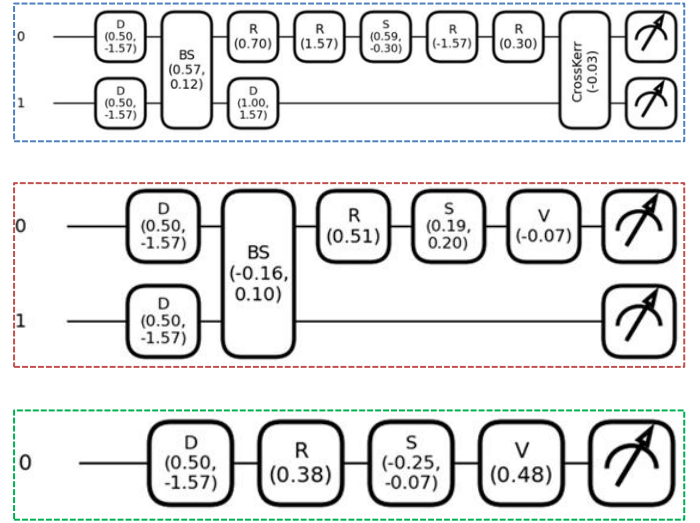

Fig. 3 The proposed CV-QNN architectures with their optimized parameters (drawn by Python). The gate names D, BS, R, S, V refer to displacement, beamsplitter, rotation, and nonlinear variable, respectively. Top: Model-1, Middle: Model-2, Bottom: Model-3.

Table I The optimized parameters and cost error for the three models

| Parameter | Model 1 | Model 2 | Model 3 |
|---|---|---|---|
| 1 | 0.56999711 | -0.15734424 | 0.38387328 |
| 2 | 0.11908 | 0.10423003 | -0.25113897 |
| 3 | 0.31154198 | 0.51234233 | -0.07175718 |
| 4 | 0.70493579 | 0.18854571 | 0.4849161 |
| 5 | 0.58592745 | 0.20172705 | - |
| 6 | -0.30231739 | -0.07449027 | - |
| 7 | 0.29977448 | - | - |
| 8 | -0.03426343 | - | - |
| MSE error | 0.0014903 | 0.0015005 | 0.0014855 |

Notably, the model achieved a loss of 0.15% and a root mean squared error of 73.11 MW—approximately a 2.6% error relative to a maximum demand of around 2750 MW. The model effectively captures underlying data patterns due to its highly nonlinear structure, the quantum entanglement between the two wires, and the superposition of the 12 Fock states used during training. Compared to the classical OTSAF model from [29], the proposed model achieved a relatively lower error (2.6% vs. 3.08%) while utilizing a much simpler network architecture with only six to eight tunable parameters and a single layer. In contrast, the OTSAF network is significantly more complex, incorporating three LSTM layers with 128 hidden units each to achieve a similar performance level. Achieving such accuracy with minimal training and no classical components underscores the power and promise of quantum architectures, laying the foundation for further advancements in sequential forecasting tasks.

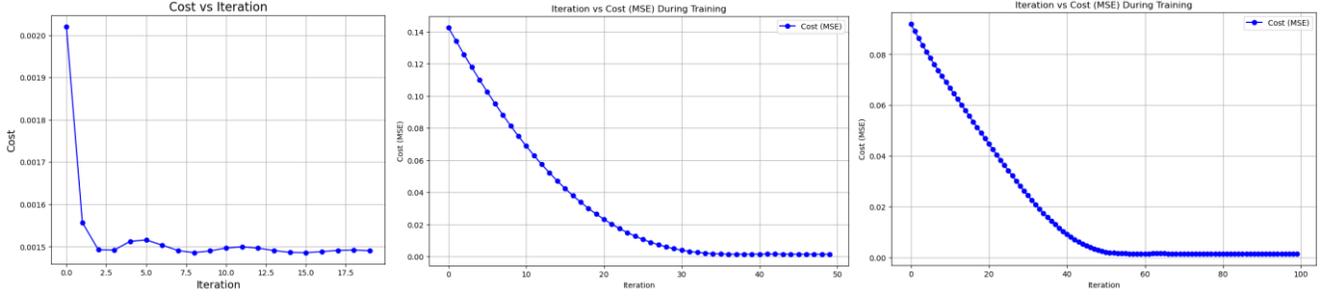

Fig. 4 Training results: error vs iteration for the three models, respectively

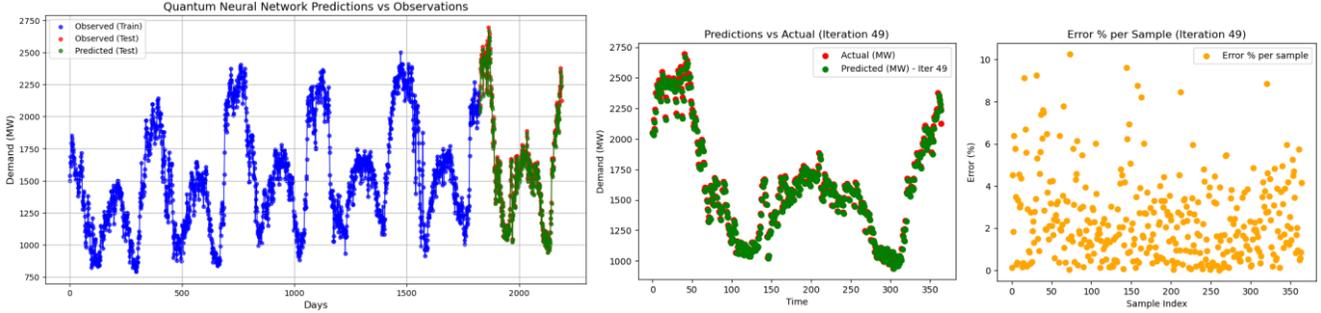

Fig. 5 The results obtained from Model- 2. The left plot illustrates the training data, test data, and predicted test data (six years in total). The middle plot compares actual vs. predicted test data (last year), while the right plot displays the error vs. sample distribution.

Figure 4 shows the error plots for the three models, whereas Figure 5 showcases the results from Model 2 revealing a gradual learning and strong matching at the end between the predictions and actual values. Other results are omitted here due to space constraints but are available with the provided codes. The left plot displays the training, test, and predicted test data spanning six years. The middle plot presents a comparison between actual and predicted test data for the final year, while the right plot illustrates the error distribution across samples. In this plot, green represents the predicted values, while red/orange indicates the actual test data .

The presented result considered the case where only load demand is to be forecasted. Next, the model is tested with a multi-feature input, incorporating both load demand and weather data (temperature in this study). Each feature is encoded onto a different wire in the quantum circuit: load demand feature is encoded on wire 0 while temperature feature is encoded on wire 1. The model is retrained for 50 epochs using the simplifications in Model 2, yielding similar cost results (0.0014888). However, the learning process is notably faster, as observed in the cost error vs. iteration plot. This improvement in training speed is likely due to the correlation between Input 1 and Input 2, where higher temperatures in hot seasons lead to increased demand, and a similar trend is observed during colder weather. The updated parameters after training are (0.08623425, 0.63306879, 0.20919965, 0.55718403, 0.6897581, -0.31122784).

Another case study is conducted using two input features: the actual demand and its derivative encoded on the wires individually. The resulting cost of 0.0014827 is comparable to that of the case with demand and temperature. Notably, the tunable parameters in both multi-feature scenarios have similar values with slightly alteration, highlighting consistency across double-input feature cases.

To investigate the proposed model on more general cases, Model Simplification 2 is extended to include 3 inputs: minimum, maximum, and mean load demands over the six years. Input 1, representing the average demand, is encoded on wire 0, while the maximum and minimum demand features are encoded on wires 1 and 2, respectively. Two beam splitters are used—one between Input 1 and Input 2, and another between Inputs 1 and 2 (requiring a number of beam splitter gates equal to the number of inputs or features minus one). The results indicate similar high performance, though the model takes more time to train with the additional complexity. The 3-wire model with the optimized parameters is shown in Fig. 6. The results are similar to those obtained from previous models, so they are not presented here. It should be noted that the beamsplitter gates are set to have identical values, though they can be configured with unique values for each. Additionally, the second beamsplitter is connected only to wires 0 and 2, as demonstrated in the highlighted section of the input-output diagram.

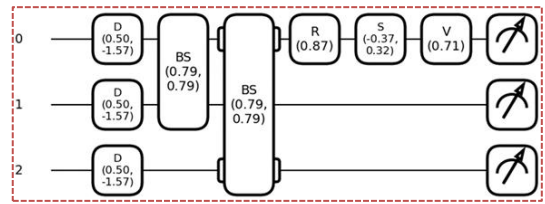

Fig. 6 Experiment of the 3-wire extended Model-2.

A more advanced forecasting scenario involves predicting demand over multiple future time steps, rather than just one step ahead as done in earlier cases. Here, the proposed Model-2 is used to predict demand for two year into the future, using the same training dataset. A function is created to predict future demand using a sliding window approach. This function takes the most recent window-size days of data as input, predicts the demand for the next day using the trained model, adds the predicted value to the window, and removes the oldest value. This process is repeated for the desired number of future days. This method allows the model to generate multi-step forecasts iteratively, where each predicted value is used as input for the next prediction. This is known as closed-loop or recursive prediction. To make the model more robust and better at handling unseen data, random noise is added to the training data, which helps improve its generalization ability.

In this experiment, a 3-year input window is chosen to emphasize recent data and seasonal patterns and reducing computational complexity. The sliding window approach is then applied iteratively to generate predictions for each day of the two-year period, with each predicted value becoming part of the input for the next prediction. This problem required extended training time. An early stopping mechanism was implemented, terminating training at epoch 20 out of a possible 50 to optimize performance. The results, shown in Fig. 7, include a blue scatter plot representing the test data from the final year of the dataset, along with the model's corresponding predictions. The model then extends its predictions an additional 365 days into the future (Year 7), successfully capturing the dataset's seasonal patterns. The predicted trend maintains the distinctive "W" shape, reflecting the region's four-season cycle. Unlike the method in [29], which only predicts test data (year 6), this model provides a full year of predictions beyond the test range. The proposed method achieves a mean absolute error (MAE) of 6.3% (171 MW out of 2696 MW), which is better than the 7.99% error reported in [29] using a deep learning approach that is limited to the test dataset and does not extend predictions beyond the six-year mark. This result highlights the effectiveness of the proposed one-layer model, which, despite its simplicity and fewer iterations, performs better than more complex models like LSTM and GRU in accurately capturing the dataset's patterns.

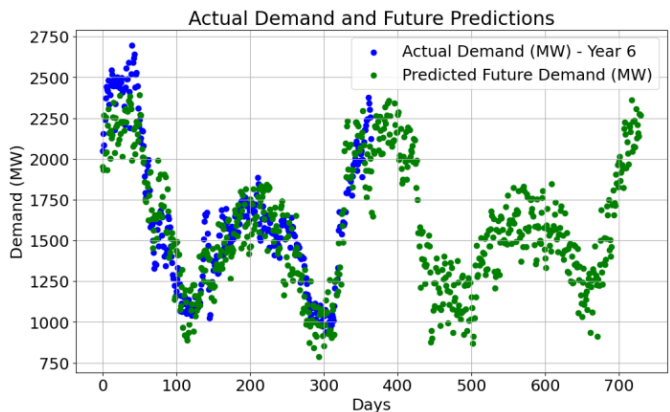

Fig. 7 Multiple time-step ahead forecasting for the following two years

## 1.5 Transfer Learning

In this section, the proposed simplification (Model-2) is utilized to predict future outcomes for various forecasting problems using the pretrained parameters saved in the last section (six parameters listed in Table 1). The study aims to apply the model to several different problem domains including large energy consumption dataset, traffic flow, vehicle count, sunspot number dataset, and cryptography. The datasets used are publicly available and are listed in Appendix I. All code implementations are provided alongside this paper, and while only selected results are presented here, the full results are included with the code and are accompanied by a demonstration video.

### 1.5.1 Dataset-2 (Hourly Forecasting Dataset)

Similar to the load demand dataset for the Kurdistan region, an hourly electricity consumption and production dataset obtained from Kaggle [30] is utilized. This dataset comprises 46,011 samples, which is approximately 25 times larger than the previous dataset. It is divided into an 80% training set and a 20% testing set. The pre-trained Model-2 from Section 1.4 is loaded and applied to the data to predict energy consumption without further training. For this task, the model achieves descent performance, with a cost of 0.0033 and percentage RMSE of 4.62%. The performance can be further improved by training the network initializing from the pretrained weights obtained in Table 1. The test predictions obtained from running Model-2 with the pretrained parameters without retraining the network are shown in Fig. 8. The results demonstrate the model's capability to generalize to new forecasting problems and function effectively as a transfer learning model.

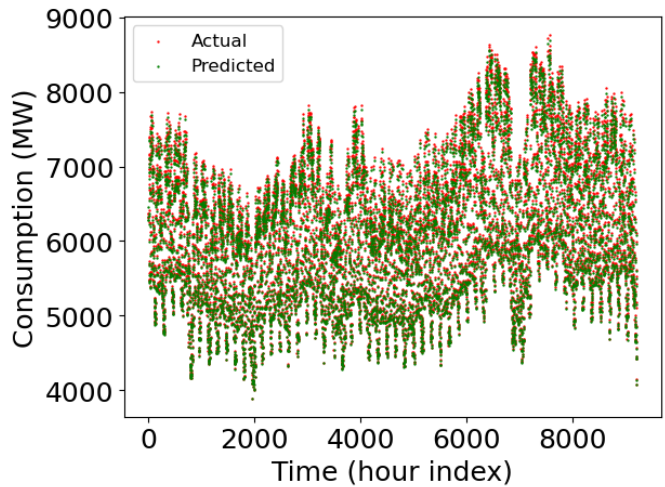

Fig. 8 Applying transfer-learning network to dataset-2

### 1.5.2 Datset-3 (Large Hourly Energy Consumption)

The third dataset, also publicly available on Kaggle, contains a larger dataset of 121,273 samples representing hourly energy consumption provided by the American Energy Power (AEP) [31]. The goal is to evaluate the model's scalability with large-scale datasets. Similarly, the pretrained weights from section 1.4 is loaded and used as transfer learning for the second simplification (Mode-2) and the results obtained shows an error of 3.82% within around a minute. This demonstrates the model's effectiveness in predicting large-scale dataset and works as a transfer learning network. The predictions for the test dataset are plotted in Figs. 9 showing that the model effectively captures the patterns in the historical data.

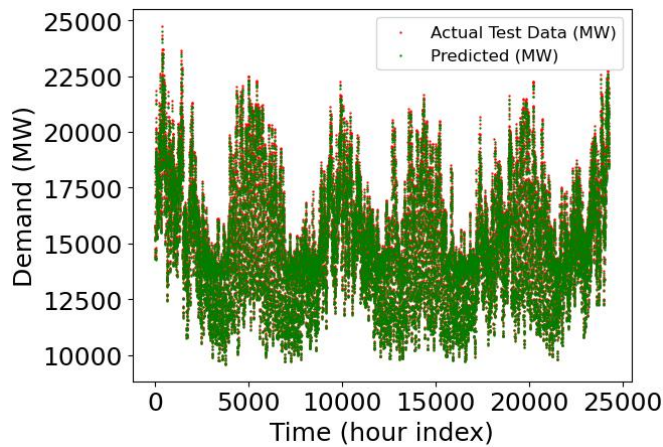

Fig. 9 Applying transfer-learning network to dataset-3

**1.5.3 Dataset 4 (Traffic Vehicle Prediction)**
To validate the effectiveness of the proposed model, a different forecasting problem is employed: Traffic Vehicle Prediction [32]. This dataset contains a total of 48,120 samples across three junctions. For this problem, predictions are made for the vehicles at junction 1, which is split into 11,672 training samples and 2,919 test samples. The predictions are plotted and shown in Fig. 10.

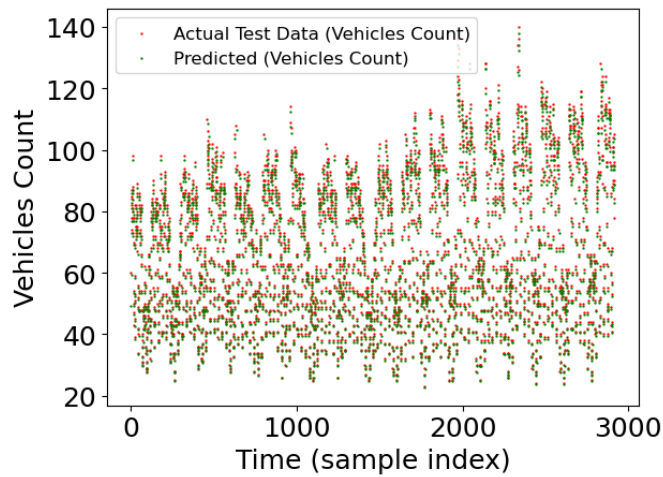

Fig. 10 Applying transfer-learning network to dataset-4

**1.5.4 Dataset 5 (Weather Prediction)**
Another forecasting problem is examined, which involves weather prediction, with a focus on the temperature feature, although the model can be extended to other features provided by the source [33]. The dataset consists of 96,453 samples, with 80% used for training and 20% for testing. The predictions for the test dataset are shown in Fig. 11.

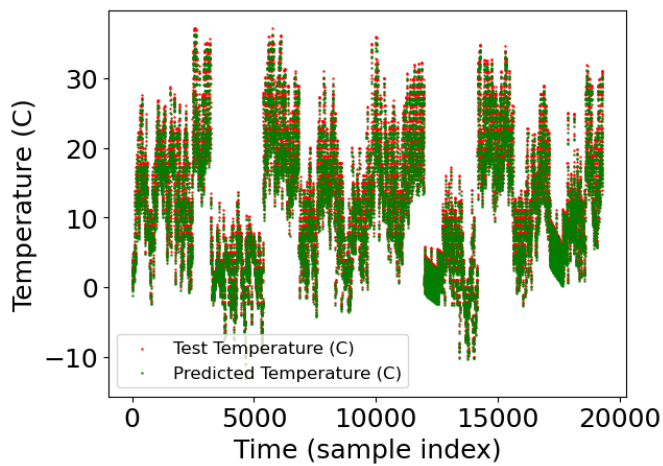

Fig. 11 Applying transfer-learning network to dataset-5

**1.5.5 Dataset 6 (Sunspot Number Prediction)**
This dataset consists of 75664 time-series samples representing sunspot numbers, which measure solar activity over time [34]. It is sourced from the SILSO (Solar Influences Data analysis Center) and contains multiple rows, with each entry corresponding to the sunspot number for a specific time interval. The dataset is employed for solar forecasting and research, with the target variable being the sunspot number in the 5th column. The results are presented in Fig. 12.

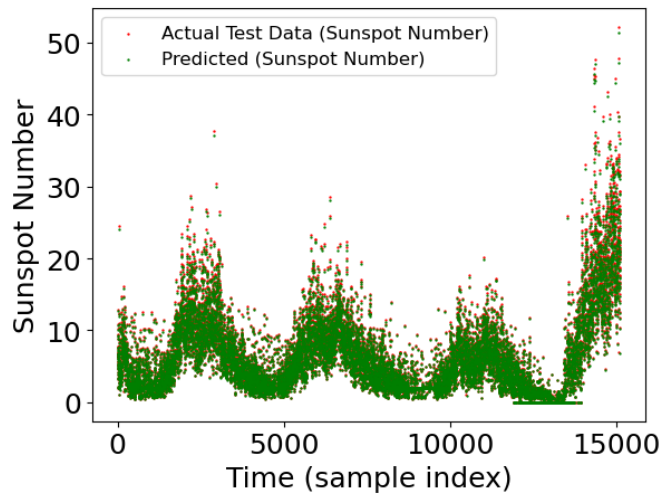

Fig. 12 Applying transfer-learning network to dataset-6

**1.5.6 Dataset 7 (Bitcoin Price Prediction)**
The Cryptocurrency Price Analysis Dataset contains daily price data for six major cryptocurrencies: Bitcoin (BTC), Ethereum (ETH), Ripple (XRP), and Litecoin (LTC). The dataset spans from January 1, 2018, to May 31, 2023, and includes columns for the cryptocurrency name, date, opening price, highest price, lowest price, and closing price for each day [35]. This dataset allows users to analyze price trends, volatility, market dynamics, and performance, with potential use cases including price analysis, volatility studies, trading strategies, and sentiment analysis. It offers insights into the long-term behavior of cryptocurrencies, useful for research, analysis, and developing models within the cryptocurrency market. The predictions from running Model-2 with the pretrained parameters are shown in Fig. 13.

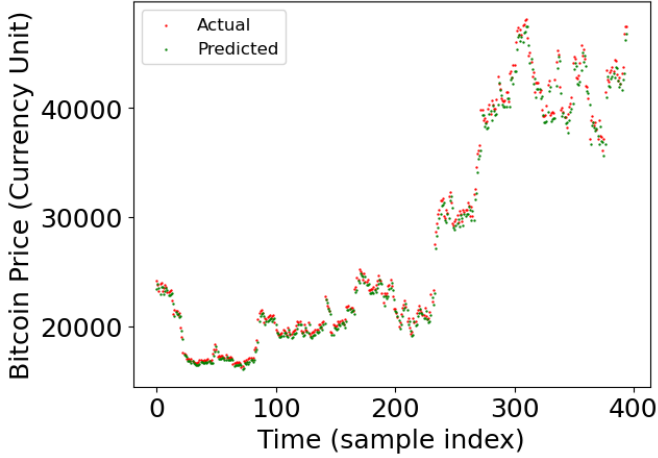

Fig. 13 Applying transfer-learning network to dataset-7

**1.6 Discrete Qubit-Based Approach**
In Section 1.2, a simple yet effective continuous-variable quantum neural network was proposed for solving forecasting problems. This section examines an alternative quantum approach using discrete variables (DV), which mirrors the configuration of the CV model while incorporating popular discrete quantum gates such as RX, RZ, RY, and CNOT, as well as encoding techniques such as angle and amplitude embedding. A two-wire network is employed, utilizing Hadamard and CNOT gates to establish entanglement. On the first wire, an RZ gate is placed between two RY gates, forming a "sandwiched" structure. One RY gate is set to a quarter-cycle angle, while the other functions as a rotation gate optimized as a weight parameter, similar to those in classical neural networks. On the second wire, an RX gate serves as a bias parameter, followed by an additional CNOT gate to introduce further entanglement, analogous to the Cross-Kerr gate. Unlike the CV models, the HOM effect is not applied to the discrete network; therefore, the second wire does not function as an ancilla or auxiliary qubit.

The proposed model is applied to forecast load demand in Kurdistan using the same train-test split ratio. The model is drawn as a diagram using Python code, with optimized parameters displayed in Fig. 14a. After training, the model is applied to predict test data, compute cost and error, and demonstrate its ability to learn and recognize hidden patterns. However, increasing the number of iterations, circuit depth, or tuning hyperparameters did not result in significant improvements. Further refinements are required to enhance predictive accuracy. Therefore, a double two-wire architecture is introduced, adding additional entanglements across all wires. This configuration doubles the wire count to four (requiring four qubits), with the third and fourth wires replicating the structure and gates used in the first two wires (Fig. 14b). A total of eight parameters are optimized with a learning rate of 0.01 and randomized initial values. Using only one layer, this approach demonstrates a fair improvement over the single two-wire structure, achieving an error of 86 MW (3.19%). However, this method requires twice the number of parameters compared to the simple two-wire version, additional entanglements, longer training times, and lower processing speed. The classical data are encoded into quantum states using two popular techniques: amplitude encoding and angle encoding. The results indicate that, for this study and the limited number of qubits, angle encoding performs significantly better than amplitude encoding. It is worth mentioning that the angle encoding used here involves rotation around the Y-axis, meaning the data are encoded by rotating them around the Y-axis. The model has also been tested as a transfer learning approach on the other datasets, performing reasonably well but with lower accuracy compared to the CV models. However, it has not been evaluated on multi-feature inputs as in the CV models, leaving this aspect for future investigation and research.

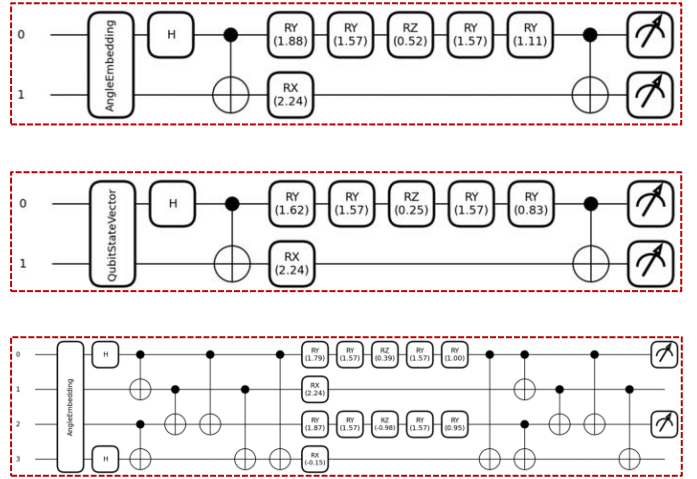

Fig. 14 The proposed DV-QNN architectures with their optimized parameters (drawn by Python). Top: two-wire QNN with angle encoding, Middle: two-wire QNN with amplitude encoding, Bottom: 4-wire QNN.

Table II Datils of the proposed DC-QNN model

| Encoding Method | No. of qubits | Cost | % Error |
| --- | --- | --- | --- |
| Amplitude Encoding | 2 | 0.0064109 | 17.71 |
| Angle Encoding | 2 | 0.0017794 | 6.13 |
| Angle Encoding | 4 | 0.0015257 | 5.35 |

**1.7 Understanding the Key Drivers of Model Success**
We may wonder why the pretrained model, with its simpler structure consisting of fewer parameters and gates, performs exceptionally well in transfer learning on unseen datasets. Despite its reduced complexity, the model retains high performance without requiring retraining on new data. One key reason for this is the specific forecasting technique used by the model. It predicts the next time step in a time series, making the output a shifted version of the input. This one-step shift is a common feature in time series forecasting tasks, and quantum neural networks (QNNs) are particularly adept at recognizing and learning such patterns due to their inherent structure and ability to model sequential relationships. Another reason lies in the seasonality of the data. Many real-world forecasting datasets, such as those related to weather or sales, exhibit seasonal patterns. The QNN can learn these patterns, allowing the model to generalize knowledge from one dataset and apply it to another. This ability is crucial for transfer learning, enabling the model to effectively transfer knowledge from one task to a related task. Additionally, QNNs operate in Hilbert space and use complex numbers, which allow them to model nonlinearity in the data. In contrast, classical neural networks rely on activation functions to introduce nonlinearity in a real-valued space. By leveraging quantum mechanical principles—particularly superposition and entanglement as foundational elements in the models—the approach achieves computational advantages inaccessible to classical systems. QNNs can capture intricate, nonlinear relationships within the data that classical networks might struggle to represent.

### 1.8 Limitations
This section presents some limitations of the study. First, while the model was evaluated across multiple datasets to assess generalization and scalability, all experiments were conducted using the Strawberry Fields simulator due to hardware access constraints. Attempts to run the model on Xanadu's X8 processor (8 qumodes) failed with the 2-qumode implementation, and access to IBM Quantum was restricted due to geographical limitations. Second, the comparative advantage of our CVQNN over DCQNN architectures remains unsubstantiated, as no systematic benchmarking was performed. Third, the model's applicability to, for instance, regression tasks was not explored, leaving its generalizability across problem domains uncertain. These gaps, along with the inherent limitations of current NISQ devices, such as noise and decoherence, highlight important boundaries to our findings. Future work should focus on hardware implementation, rigorous architectural comparisons, and extending the model to regression problems to fully validate the approach.

### 1.9 Conclusion
This study presented firstly a continuous-variable quantum neural network (CV-QNN) for forecasting problems. The proposed CV-QNN model achieves high accuracy with minimal complexity, utilizing only a single layer, two wires (two qubits), and mostly eight trainable parameters. Applied to load forecasting for the Kurdistan regional power system and two additional electricity consumption datasets, the model demonstrated its capability to capture complex patterns, achieving accuracies on par with state-of-the-art deep learning models. The study also highlights the robustness of CV-QNN in transfer learning applications, where freezing initial layers while fine-tuning to no-tuning across another seven different datasets. The model's success in multi-step forecasting and multi-feature scenarios further confirms its flexibility and accuracy, meeting a key need in sequential forecasting applications. Additionally, a discrete-variable quantum neural network (DV-QNN) approach was investigated, proposing two-wire and four-wire configurations with discrete quantum gates. Overall, the findings highlight the strengths of the CV-QNN while illustrating the potential and challenges of the DV-QNN in sequential machine learning tasks. AT the end this study, the limitations of this research and the factors contributing to the model success are presented.

**Funding**
No financial support is provided for this work.

**Conflict of Interest**
The authors declare that there is no conflict of interest associated with this work.

**Data Availability**
All datasets used in this study are publicly accessible in Appendix I.

**Code Availability**
https://github.com/IsmaelAbdulrahman/CVQNN_DCQNN/tree/e8aecd502d0f099e9a48bed3b9740f3d9a656112

**Biography**
Ismael Abdulrahman received his PhD in Electrical and Computer Engineering from Tennessee Technological University, USA, in 2019. Currently, he serves as an assistant professor at Erbil Polytechnic University, where he instructs graduate and 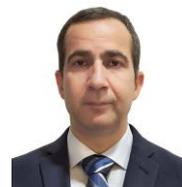 undergraduate courses including quantum computing, advanced deep learning, advanced mathematics, and electrical and electronic courses. His academic passions include quantum computing, deep learning, optimization, and mathematical modeling.

### References
1  Nielsen, Michael A., and Isaac L. Chuang. *Quantum Computation and Quantum Information*. Cambridge University Press, 2010.

**Appendix I: Dataset Sources**